\documentclass{article}
\usepackage{subcaption}
\usepackage{spconf,amsmath,amssymb,graphicx,hyperref}
\usepackage{array}
\usepackage{booktabs}
\usepackage{multirow}
\usepackage{nicematrix}
\usepackage{tikz}
\usepackage{threeparttable}
\usepackage{siunitx}
\usepackage{amssymb}

\title{ToneMamba: Learning Contextualized Tonal Contours with Boundary-Reset State Space Models}
\name{Po-Yu Chiu, Yi-Fen Liu \thanks{Code and models are available at:\\ \url{https://github.com/fcu-slp/ToneMamba}}}
\address{Department of Information Engineering and Computer Science, Feng Chia University, Taiwan}
\begin{document}
%\ninept
%
\maketitle
\begin{abstract}
Mandarin tone recognition in connected speech is challenging because contextual and durational variation reshapes F0 contours. We propose ToneMamba, a boundary-aware state space model that replaces static boundary embeddings with bidirectional state resets at syllable, word, and IPU levels. Fixed weights fuse multi-level representations, while shared-weight bidirectional scans use channel-wise directional fusion to reduce parameters. Dynamic syllable pooling forms tonal contours, and rhythm-conditioned gating incorporates durational information. ToneMamba achieves 97.9\% and 74.8\% accuracy on FCU-VOICE-360 read and MCDC-8 conversational speech, outperforming the baseline (ToneNet) by 0.9 and 15.9 percentage points, respectively. Ablations show that word-level resets are particularly effective, while rhythm conditioning provides greater gains in conversational speech, underscoring the role of durational variation in spontaneous speech.
\end{abstract}

\section{Introduction}

Fundamental frequency (F0) is widely recognized as the primary acoustic correlate of Mandarin lexical tones. Prior work \cite{huang2021encoder} showed that improved F0 tracking with an encoder-decoder framework and gating mechanism achieves competitive tone recognition performance, comparable to systems using MFCCs, FFT spectrograms, or raw waveforms \cite{yang2018bidirectional,tang2021context}. More recently, leveraging wav2vec 2.0, \cite{yuan2023contextualized} extended this line of research by employing a Transformer-based self-supervised framework to learn robust contextualized tonal contour representations.

In connected speech, a syllable's pitch trajectory is shaped by neighboring tonal contexts and durational variation \cite{lee2021pre,Xu_Lee_2022}. ToneNet \cite{liu2024contextualized}\cite{lu24d_interspeech} learns tonal contour representations from F0 sequences, syllable and word boundaries, and rhythmic information. Its Transformer-based contour encoder learns representations bottom-up from F0 and consolidates syllable-level contours through cross-attention with rhythmic contexts. Although designed embeddings encode syllable- and word-level structure, ToneNet does not explicitly constrain F0 tokens or [CLS]-based representations from attending beyond their linguistic boundaries.

The structured state space duality (SSD) formulation of Mamba-2 makes hidden-state propagation explicit \cite{dao2024transformers}, enabling boundary-dependent state resets. Leveraging this property, we propose ToneMamba, a boundary-aware state space model (SSM) for Mandarin tone recognition. Our contributions are threefold. First, hierarchical linguistic segmentations from syllable to word and IPU levels are incorporated as independent boundary-reset mechanisms in Mamba-2 rather than Transformer embeddings. Second, shared-weight bidirectional scans use channel-wise directional fusion to reduce fusion parameters. Third, cross-attention with rhythmic contexts is replaced by a gate that controls their influence before boundary-reset Mamba-2 processing.

\begin{figure}[t]
    \centering
    \includegraphics[width=\columnwidth]{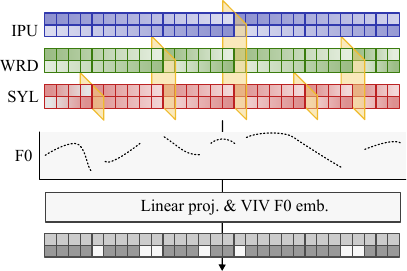}
    \caption{Top to bottom: Reset indices at IPU, word (WRD), and syllable (SYL) boundaries, controlling forward and backward SSM state resets. Projected F0 representations are combined with valid/invalid (VIV) embeddings.}
    \label{fig:model_input}
\end{figure}

\begin{figure*}[t]
    \centering
    \includegraphics[width=\textwidth]{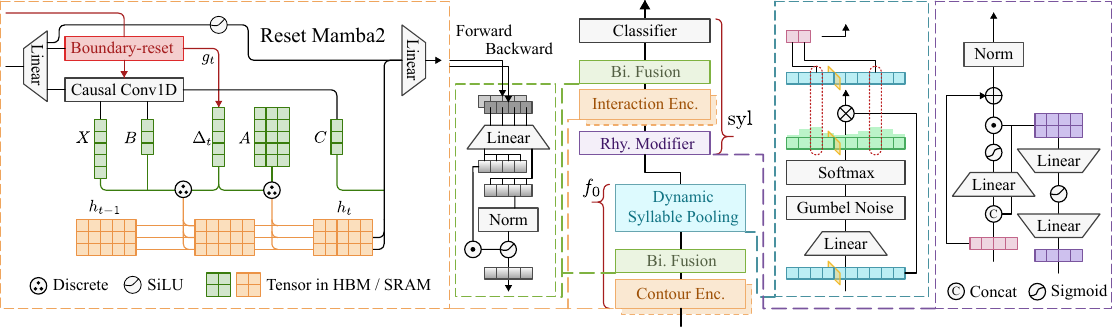}
    \caption{Overview of ToneMamba. The boundary-reset Contour Encoder (CE) encodes F0 frames. Bidirectional fusion and dynamic pooling produce syllable representations. The rhythm modifier adds duration features before the boundary-reset Interaction Encoder (IE) models their interactions for classification.}
    \label{fig:overview}
\end{figure*}

\section{Model Framework}
\label{sec:method}

\subsection{Inputs for Contour Encoder}
\label{ssec:input_representation}
The contour encoder in ToneNet \cite{liu2024contextualized,lu24d_interspeech} encodes F0 tokens with embeddings that mark aligned linguistic boundary changes and indicate each token's status within the syllable- and word-level structure of an utterance. The resulting input embeddings are defined as follows:
\begin{equation}
    x^{\mathrm{attn}}_{t}=\mathrm{LN}\left(W_f f_t+b_f+E_s(c_{t,s})+E_w(c_{t,w})+p_t\right)
    \label{eq:transformer_static_boundary}
\end{equation}
where $f_t$ is the normalized log-F0 value at frame $t$, linearly projected into a $d$-dimensional model space ($d = 256$) and combined with aligned code embeddings ($c_{t,s}$ and $c_{t,w}$) encoding each token's membership and content at the syllable and word levels, while $p_t$ denotes the positional embedding. However, Transformer attention does not explicitly constrain these embeddings or hierarchically learned representations within their corresponding linguistic boundaries. 

By exposing hidden-state propagation, the SSD formulation enables explicit control over state transitions \cite{dao2024transformers, gu2024mamba} and allows states to be reset at predefined linguistic boundaries. Building on its successful applications to speech, audio, and pronunciation training \cite{zhang2025mambaspeech,erol2024audiomamba,chao2025hierarchicalssm}, we disentangle the membership- and content-aware embeddings and recast their linguistic boundary information as explicit state-reset mechanisms. As shown in Fig.~\ref{fig:model_input}, we define:
\begin{equation}
\begin{aligned}
    e_{t,\mathrm{valid}} &= E_v(u_t)\\
    x_t &= \mathrm{RMSNorm}\left(W_f f_t+b_f+e_{t,\mathrm{valid}}\right)
\end{aligned}
    \label{eq:f0_projection}
\end{equation}
where, $e_{t,\mathrm{valid}}$ denotes valid/invalid F0 embedding which is combined with the linearly projected log-F0 before RMS normalization, yielding ($x_t\in\mathbb{R}^{d}$). During training, Gaussian noise is added to the projected log-F0 for augmentation.

\subsection{Boundary-reset in Reset Mamba2}
\label{ssec:boundary_reset_ssm}

 Boundary resetting is supported in both Mamba and Mamba-2: Mamba uses a selective SSM with a diagonal state structure, whereas Mamba-2 adopts a scalar-structured SSM based on structured state space duality (SSD) \cite{dao2024transformers}. Mamba-2 constrains $A$ to a scalar-times-identity structure, supporting Transformer-like head dimensions (e.g., ($\text{P}=64$ or $128$)), where the scalar $a_t$ is shared across channels within each head, while $B$ and $C$ can additionally be shared across heads. SSD also enables faster training than general SSMs through efficient matrix multiplications. 

Following the gated-RNN interpretation in Theorem 1 of \cite{gu2024mamba}, selective SSM recurrence can reset its state at boundaries through a gating control $g_t$. To explicitly control boundary-dependent state resets, we incorporate this gating mechanism into the reformulated SSM:
\begin{align}
    h_t &= A_t\left(g_t h_{t-1}\right)+B_t x_t, \label{eq:ssm_state}\\
    y_t &= C_t^\top h_t \label{eq:ssm_output}
\end{align}
where $h_t$ denotes the contour state memory and $A_t$ controls the decay of F0 information retained in the hidden state.

For each linguistic level $m$, i.e., syllable, word, or IPU, let $s_m(t)$ denote the index of the linguistic unit containing frame $t$, with a boundary reset triggered when $s_m(t) \neq s_m(t-1)$, as defined below. Replacing $g_t$ in \eqref{eq:ssm_state} with $g_{m,t}$ resets the previous state when $r_{m,t}=1$, allowing each linguistic unit to start from a fresh state.
\begin{equation}
\begin{aligned}
    r_{m,t} &= \mathbf{1}\left[t=1\ \mathrm{or}\ s_m(t)\ne s_m(t-1)\right]\\
    g_{m,t} &= 1-r_{m,t}
\end{aligned}
    \label{eq:reset_keep_gate}
\end{equation}

\subsection{Bidirectional and Multi-level Linguistic Fusion}
\label{ssec:bidirectional_reset_fusion}
The matrix transformation of SSMs exhibits an explicit semiseparable structure along the sequence dimension $\text{T}$ and is applied to F0 sequence inputs $X$ in both forward and backward directions, independently at each linguistic-level $m$. Notably, the $+$ and $-$ superscripts denote scan directions rather than separate parameter sets. 
\begin{equation}
\begin{aligned}
    H^{m,+} &= M^{m,+}X,
    & M^{m,+}_{ji} &= Q^{m,+}_{ji}C_j^\top A_j\cdots A_{i+1}B_i,\\
    H^{m,-} &= M^{m,-}X, 
    & M^{m,-}_{ji} &= Q^{m,-}_{ji}C_j^\top A_j\cdots A_{i-1}B_i,
\end{aligned}
    \label{eq:bidirectional_outputs}
\end{equation}

$Q$ in \eqref{eq:bidirectional_outputs} and \eqref{eq:bidirectional_support} is a particular representation of the identity matrices, indicating that the current F0 token shares the same linguistic unit with preceding F0 tokens in the forward direction and subsequent F0 tokens in the backward direction. The backward branch reverses the input sequence and boundary indices, applies the shared encoder, and then restores the original order.
\begin{equation}
\begin{aligned}
    Q^{m,+}_{ji} &= \mathbf{1}[i\le j]\mathbf{1}[s_m(j)=s_m(i)]\\
    Q^{m,-}_{ji} &= \mathbf{1}[i\ge j]\mathbf{1}[s_m(j)=s_m(i)]
\end{aligned}
    \label{eq:bidirectional_support}
\end{equation}

To reduce fusion parameters, the bidirectional hidden states at each time step $t$ are fused channel-wise as follows:
\begin{equation}
\begin{aligned}
    \left[u_{m,c}\;\;v_{m,c}\right]
    &=\frac{1}{\sqrt{2}}\left[h_{m,c}^+\;\;h_{m,c}^-\right]W_c+b_c\\
    W_c&\in\mathbb{R}^{2\times 2}
\end{aligned}
    \label{eq:dirmix_projection}
\end{equation}

\begin{table*}[!t]
\centering
\caption{Boundary-reset ablation across linguistic levels. ToneNet uses static boundary embeddings, whereas ToneMamba uses boundary-dependent state resets. $\vartriangle$ and $\blacktriangle$ denote the linguistic levels applied in the CE and IE, respectively; $+\texttt{r}$ indicates the rhythm modifier.}
\label{tab:ablation}
\vspace{4pt}
\small
\renewcommand{\arraystretch}{0.90}
\setlength{\tabcolsep}{3pt}
\begin{NiceTabular}{
    w{l}{16em}@{\hspace{24pt}}
    *{2}{w{c}{3.0em}}@{\hspace{10pt}}
    *{3}{w{c}{4.0em}}
    w{c}{4.5em}
}
\toprule
\multicolumn{1}{c}{}
& \multicolumn{2}{c}{Acc. (\%)}
& \multicolumn{3}
  {@{\hspace{8pt}}c@{\hspace{12pt}}}
  {Linguistic Level (CE $\vartriangle$ / IE $\blacktriangle$)}

& \\

\cmidrule(r){2-3}
\cmidrule(lr){4-6}

\textbf{Model}$_{(\texttt{CE} \to \texttt{IE})}$
& FCU
& MCDC
& SYL
& WRD
& IPU
& Params (M) \\

\midrule

\multicolumn{7}{l}{\textbf{Static boundary-embedding with $\text{PE}_{\text{rel}}$}} \\
\addlinespace[2pt]

% TNet-Full \cite{lu24d_interspeech} & 97.0 & 58.9 & $\vartriangle$ & $\vartriangle \mkern -6mu \blacktriangle$ & $\vartriangle \mkern -6mu \blacktriangle$ \\
\hspace{1em} $\text{ToneNet}^{\texttt{+r}}$ (a.k.a. TNet-Full in \cite{liu2024contextualized,lu24d_interspeech}) & 97.0 & 58.9 & $\vartriangle$ & $\vartriangle \mkern -6mu \blacktriangle$ & $\vartriangle \mkern -6mu \blacktriangle$ & 6.9 \\

\specialrule{0.45pt}{1.75pt}{0.4pt}
\specialrule{0.45pt}{0pt}{1.75pt}

\multicolumn{7}{l}{\textbf{Boundary-reset variants (ToneMamba)}} \\
\addlinespace[2pt]

% \multicolumn{6}{l}{\hspace{0.6em}Single-level CE} \\
\hspace{1em} $\text{ToneM}_{\texttt{s} \to \texttt{u}}$ & 93.0 & 61.7 & $\vartriangle$ & & $\mkern 12mu \blacktriangle$ & -- \\
\hspace{1em} $\text{ToneM}_{\texttt{w} \to \texttt{u}}$ & \textbf{96.2} & \textbf{63.3} & & $\mkern -12mu \vartriangle$ & $\mkern 12mu \blacktriangle$ & 5.3 \\
\hspace{1em} $\text{ToneM}_{\texttt{u} \to \texttt{u}}$ & 94.4 & 60.2 & & & $\vartriangle \mkern -6mu \blacktriangle$ & -- \\
\hspace{1em} $\text{ToneM}_{\texttt{u} \to \texttt{u}}^{\texttt{+r}}$ & \underline{\textbf{97.5}} & \underline{\textbf{65.1}} & & & $\vartriangle \mkern -6mu \blacktriangle$ & 5.4 \\

\midrule

% \multicolumn{6}{l}{\hspace{0.6em}Multi-level CE} \\
\hspace{1em} $\text{ToneM}_{\texttt{sw} \to \texttt{u}}$ & 96.2 & 63.8 & $\vartriangle$ & $\mkern -12mu \vartriangle$ & $\mkern 12mu \blacktriangle$ & -- \\
\hspace{1em} $\text{ToneM}_{\texttt{su} \to \texttt{u}}$ & 95.2 & 62.1 & $\vartriangle$ & & $\vartriangle \mkern -6mu \blacktriangle$ & -- \\
\hspace{1em} $\text{ToneM}_{\texttt{wu} \to \texttt{u}}$  & 96.6 & 64.2 & & $\mkern -12mu \vartriangle$ & $\vartriangle \mkern -6mu \blacktriangle$ & -- \\
\hspace{1em} $\text{ToneM}_{\texttt{swu} \to \texttt{u}}$ & \textbf{96.6} & \textbf{65.1} & $\vartriangle$ & $\mkern -12mu \vartriangle$ & $\vartriangle \mkern -6mu \blacktriangle$ & 5.3 \\
\hspace{1em} $\text{ToneM}_{\texttt{swu} \to \texttt{u}}^{\texttt{+r}}$ & \underline{\textbf{97.9}} & \underline{\textbf{72.9}} & $\vartriangle$ & $\mkern -12mu \vartriangle$ & $\vartriangle \mkern -6mu \blacktriangle$ & 5.4   \\

\midrule

% \multirow[t]{2}{*}{\hspace{0.6em}Word-level IE}
\hspace{1em} $\text{ToneM}_{\texttt{wu} \to \texttt{wu}}$ & 96.4 & 66.0 & & $\vartriangle \mkern -6mu \blacktriangle$ & $\vartriangle \mkern -6mu \blacktriangle$ & -- \\
\hspace{1em} $\text{ToneM}_{\texttt{swu} \to \texttt{wu}}$& \textbf{96.5} & \textbf{67.0} & $\vartriangle$ & $\vartriangle \mkern -6mu \blacktriangle$ & $\vartriangle \mkern -6mu \blacktriangle$ & 5.3 \\
\hspace{1em} $\text{ToneM}_{\texttt{swu} \to \texttt{wu}}^{\texttt{+r}}$ & \underline{\textbf{97.9}} & \underline{\textbf{74.8}} & $\vartriangle$ & $\vartriangle \mkern -6mu \blacktriangle$ & $\vartriangle \mkern -6mu \blacktriangle$ & 5.4 \\

\bottomrule
\CodeAfter
\tikz {
    \draw[line width=0.4pt, shorten <=3pt, shorten >=3pt] (4-|4) -- (5-|4);
    \draw[line width=0.4pt, shorten <=3pt, shorten >=3pt] (5-|4) -- (10-|4);
    \draw[line width=0.4pt, shorten <=3pt, shorten >=3pt] (10-|4) -- (15-|4);
    \draw[line width=0.4pt, shorten <=3pt, shorten >=3pt] (15-|4) -- (18-|4);
    
    \draw[line width=0.4pt, shorten <=3pt, shorten >=3pt] (4-|7) -- (5-|7);
    \draw[line width=0.4pt, shorten <=3pt, shorten >=3pt] (5-|7) -- (10-|7);
    \draw[line width=0.4pt, shorten <=3pt, shorten >=3pt] (10-|7) -- (15-|7);
    \draw[line width=0.4pt, shorten <=3pt, shorten >=3pt] (15-|7) -- (18-|7);
}
\end{NiceTabular}
\end{table*}

The channel-wise fused representations $U_m$ and $V_m$ are furhter integrated through multiplicative gating, where the normalized and SiLU-activated $V_m$ modulates $U_m$, yielding the linguistic-level representation $Z_m$:
\begin{equation}
    Z_m=U_m\odot\mathrm{SiLU}(\mathrm{RMSNorm}(V_m))
    \label{eq:direction_fusion}
\end{equation}

Bidirectional fusion is applied independently to each linguistic level $m$ in $G=\{\mathrm{IPU},\mathrm{word},\mathrm{syllable}\}$. We adopt a fixed logits $(\alpha_{\mathrm{IPU}},\alpha_{\mathrm{WRD}},\alpha_{\mathrm{SYL}})=(1,2,3)$ to progressively emphasize representations from IPU to syllable level. Softmax normalization yields the corresponding fusion weights $(0.090,0.245,0.665)$, which are used to obtain the final fused representation $Z$:
\begin{equation}
\begin{aligned}
    w_m &= \frac{\exp(\alpha_m)}{\sum_{n\in G}\exp(\alpha_n)},\qquad Z &= \sum_{m\in G}w_m Z_m
\end{aligned}
    \label{eq:multilevel_contour_fusion}
\end{equation}

\subsection{Dynamic Syllable Pooling}
\label{ssec:dynamic_syllable_pooling}

A tonal contour is perceived over the entire syllable span. Unlike the Transformer-based framework in \cite{liu2024contextualized}, which summarizes each syllable contour using a predefined [CLS] token, we employ hard Gumbel-Softmax to dynamically select salient F0 representations within each syllable. For the $i$-th syllable, $s_i$ and $e_i$ denote the indices of its first and last F0 tokens, respectively. Gumbel perturbations \cite{baevski2020wav2vec} are added to the selection logits $q_t$, and the resulting syllable-internal probabilities $\widetilde p_{i,t}$ are used to pool the frame-evel representations $Z_t$ into the syllable-level contour representation $c_i$:
% \begin{equation}
% \begin{aligned}
%    g_{i,t} &= -\log[-\log(u_{i,t})],\quad u_{i,t}\sim\mathcal{U}(0,1)\\
%    \widetilde p_{i,t} &= \frac{\exp((q_t+g_{i,t})/\tau)}{\sum_{k=s_i}^{e_i}\exp((q_k+g_{i,k})/\tau)}\\
%    c_i &= \sum_{t=s_i}^{e_i}p_{i,t}Z_t
% \end{aligned}
    \label{eq:dynamic_syllable_pooling_ops}
% \end{equation}
\begin{equation}
\begin{aligned}
    g_{i,t} &=
    \begin{cases}
        -\log[-\log(u_{i,t})],\quad u_{i,t}\sim\mathcal{U}(0,1),
        & \text{training},\\
        0,
        & \text{inference},
    \end{cases}\\
    \widetilde p_{i,t} &=
    \frac{\exp((q_t+g_{i,t})/\tau)}
    {\sum_{k=s_i}^{e_i}\exp((q_k+g_{i,k})/\tau)}\\
    c_i &= \sum_{t=s_i}^{e_i} \widetilde p_{i,t}Z_t
\end{aligned}
\label{eq:dynamic_syllable_pooling_ops}
\end{equation}
where $q_t=\textbf{w}_p^\top Z_t+b_p$ denotes the selection logit, with $\textbf{w}_p \in \mathbb{R}^d$, $g_{i,t}$ is the sampled Gumbel noise, and the temperature is fixed at $\tau=1$. During training, hard selection is performed using the one-hot argmax of $\widetilde p_{i,t}$, with gradients propagated through the straight-through estimator. At inference, the Gumbel noise is removed and the selection weights are obtained by applying softmax directly to $q_t$. The resulting syllable-level representations $C=(c_1,\ldots,c_{N_s})$ are subsequently passed to the Reset Mamba2-based Interaction Encoder (IE), where bidirectional multi-level linguistic fusion further refines the tonal contours, as shown in Fig.~\ref{fig:overview}.

\subsection{Rhythm Modifier}
\label{ssec:rhythm_modifier}

To model durational variability \cite{dellwo2015rhythmic}, we replace Transformer-based cross-attention with a rhythm modifier that adaptively integrates rhythmic information into each syllable-level contour representation through gated modulation. For syllable $i$, the duration vector $d_i$ concatenates the same durational measures as in \cite{liu2024contextualized,lu24d_interspeech}, computed over the whole syllable, vocalic final, and vowel intervals :
\begin{equation}
    d_i=\left[d_i^{\mathrm{syll}};d_i^{\mathrm{final}};d_i^{\mathrm{vowel}}\right]
    \label{eq:duration_features}
\end{equation}

The duration vector is projected into a rhythmic representation $v_i$, which is jointly conditioned on the contour representation $c_i$ to derive a gating vector $g_i$. The gate adaptively controls the contribution of $v_i$ to $c_i$, yielding the rhythm-modified contour representation $\tilde{c}_i$:
\begin{align}
    v_i &= W_v\,\mathrm{SiLU}(W_d d_i+b_d)+b_v
    \label{eq:duration_embedding}\\
    g_i &= \mathrm{sigmoid}\!\left(W_g[c_i;v_i]+b_g\right)
    \label{eq:duration_gate}\\
    \tilde{c}_i &= \mathrm{LN}\left(c_i+g_i\odot v_i\right)
    \label{eq:rhythm_modified_syllable}
\end{align}
forming the rhythm-modified sequence $\tilde{C}=(\tilde{c}_1,\ldots,\tilde{c}_{N_s})$.

\section{Experiments and Results}
\label{sec:experiments}

\subsection{Datasets and Model Setup}

The extended, genre-balanced read speech dataset FCU-VOICE-360 and the conversational speech dataset MCDC-8 are used to train and evaluate both the ToneNet baseline \cite{liu2024contextualized,lu24d_interspeech} and the proposed ToneMamba. ToneMamba uses a 256-dimensional model with eight CE and four IE Mamba-2 layers, each with an SSM state size of 64. The classifier consists of a 256-to-256 $\tanh$ projection, 0.3 dropout, and a 256-to-5 output layer for classifying the five Mandarin lexical tones. The model is trained on FCU-VOICE-360 for 384 epochs and fine-tuned on MCDC-8 for another 384 epochs from the best FCU checkpoint, using cross-entropy loss and AdamW \cite{loshchilov2017decoupled} with an initial learning rate of $5\times10^{-3}$, cosine decay, and 10\% warm-up. The best development checkpoint is used for test evaluation.

\begin{figure}[t]
    \centering

    \begin{subfigure}{0.48\linewidth}
        \centering
        \includegraphics[width=\linewidth]{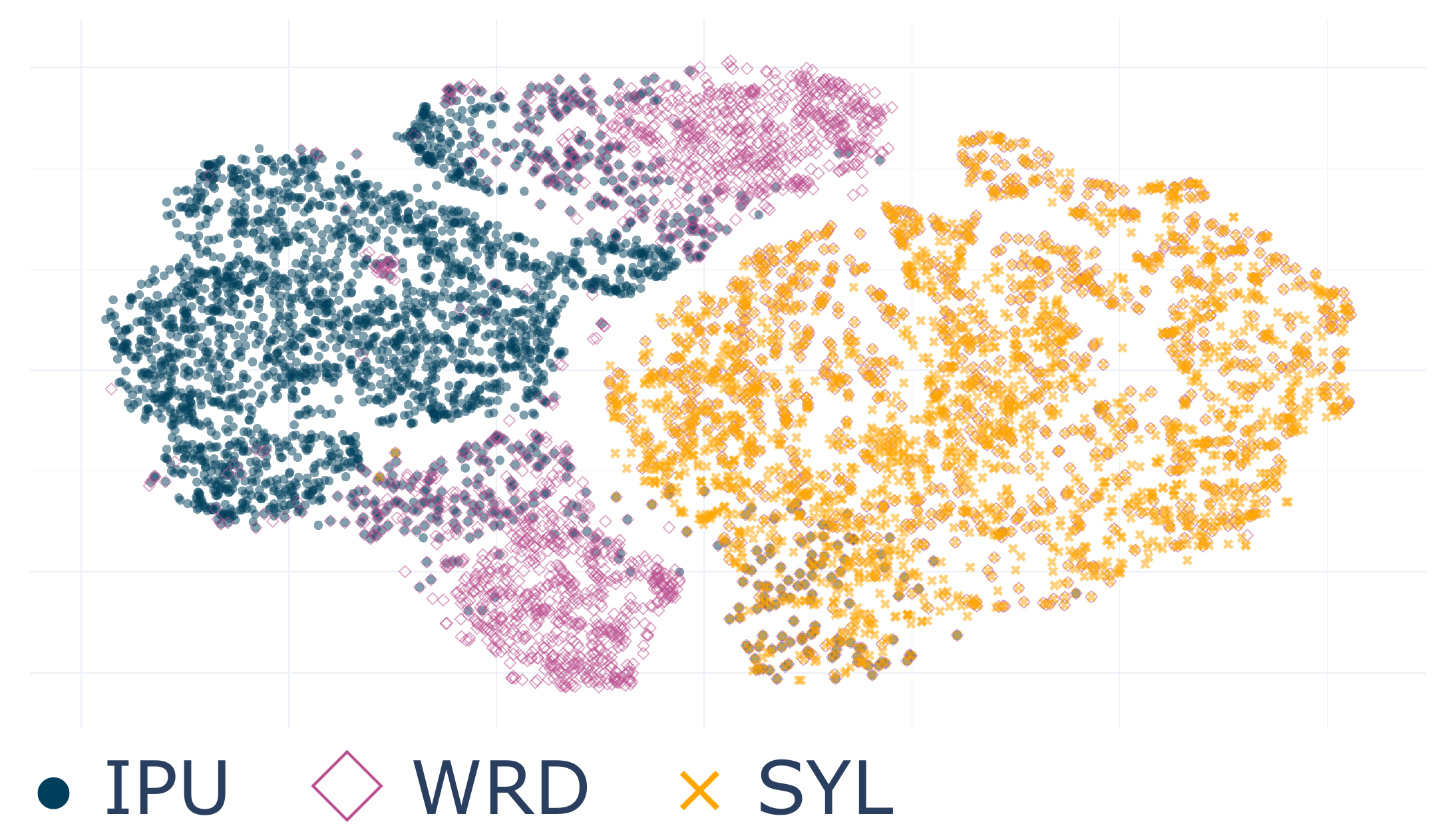}
        \caption{FCU-VOICE-360}
        \label{fig:fcu_tsne}
    \end{subfigure}
    \hfill
    \begin{subfigure}{0.48\linewidth}
        \centering
        \includegraphics[width=\linewidth]{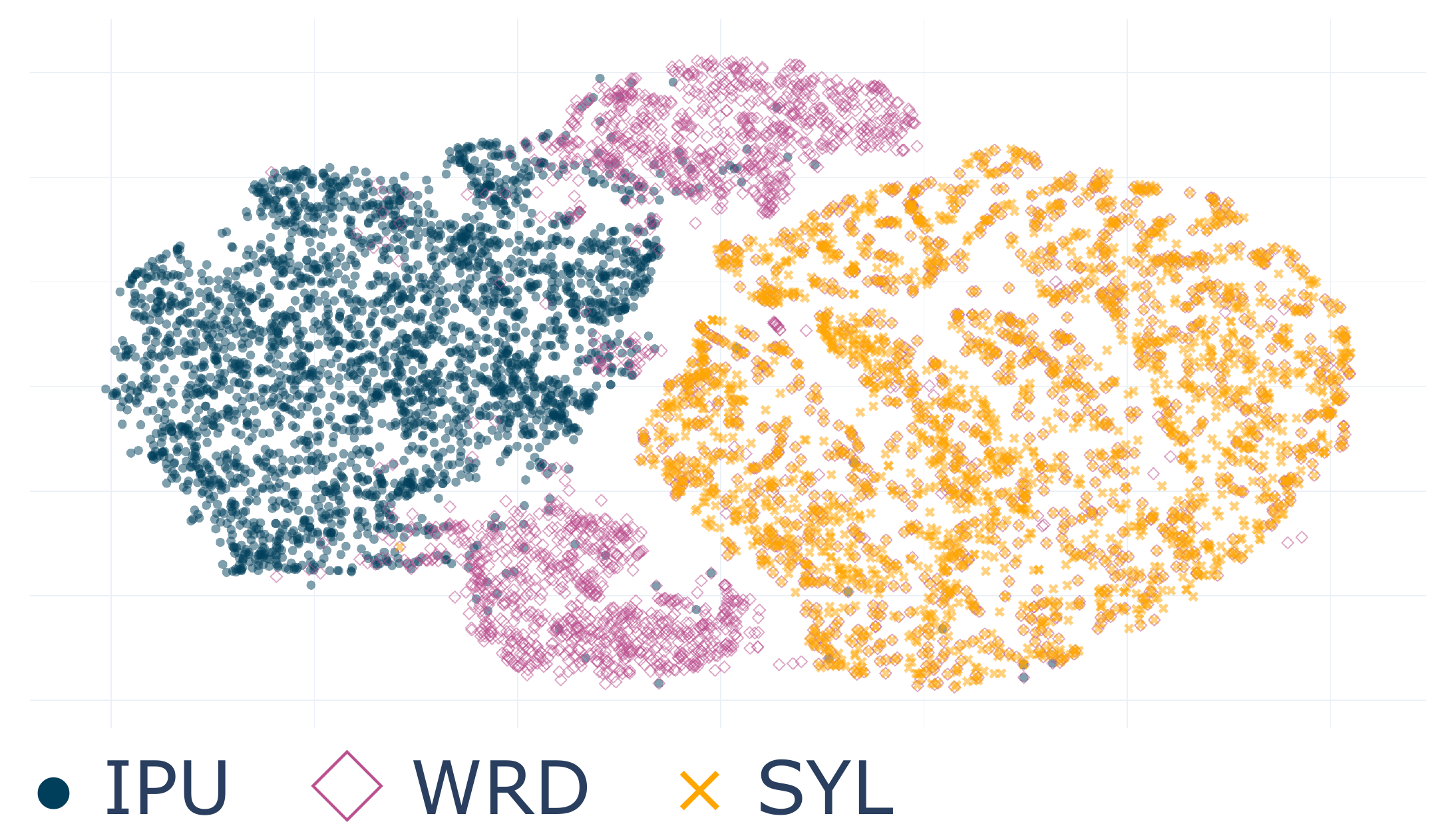}
        \caption{MCDC-8}
        \label{fig:mcdc8_tsne}
    \end{subfigure}

    \caption{t-SNE visualizations of $Z_{\mathrm{IPU}}$, $Z_{\mathrm{WRD}}$, and $Z_{\mathrm{SYL}}$, with each point representing the mean bidirectionally fused F0 representation over a valid syllable.}
    \label{fig:tsne}
\end{figure}

\subsection{Ablation Studies}

Unlike Transformer-based ToneNet, which encodes linguistic boundaries through static embeddings and relative positions, ToneMamba explicitly uses boundary-dependent state resets in both encoders. As shown in Table~\ref{tab:ablation}, under same boundary configuration, $\text{ToneM}_{\texttt{swu} \to \texttt{wu}}^{\texttt{+r}}$ further replaces cross-attention to rhythmic information with the gated rhythm modifier, achieving 97.9\% on read speech while substantially improving conversational accuracy from 58.9\% to 74.8\%, indicating greater robustness to its increased durational variability.

Without the rhythm modifier and with the IE restricted to IPU-level resets, $\text{ToneM}_{\texttt{wu}\to\texttt{u}}$ outperforms the $\text{ToneM}_{\texttt{sw}\to\texttt{u}}$ and $\text{ToneM}_{\texttt{su}\to\texttt{u}}$ configurations across both speaking styles, highlighting the importance of word- and IPU-level boundaries. Adding syllable-level resets maintains 96.6\% read-speech accuracy while improving conversational accuracy from 64.2\% to 65.1\%. The rhythm modifier further raises it to 72.9\%, while adding word-level IE resets increases it to 74.8\%, with read-speech accuracy preserved at 97.9\%. 

We further evaluate single-level CE boundary resets while restricting the IE to IPU-level resets. Word-level resets perform best, with $\text{ToneM}_{\texttt{w}\to\texttt{u}}$ achieving 96.2\% and 63.3\% on read and conversational speech, respectively, compared with 93.0\%/61.7\% and 94.4\%/60.2\% for syllable- and IPU-level resets. Adding the rhythm modifier to $\text{ToneM}_{\texttt{u} \to \texttt{u}}^{\texttt{+r}}$ further improves accuracy to 97.5\%/65.1\%, confirming the complementary benefit of durational information. 

Overall, the rhythm modifier consistently improves recognition across reset configurations. To further examine the learned representations, Figure~\ref{fig:tsne} visualizes syllable-wise averages of bidirectionally fused frame-level F0 representations $Z_m$ from $\text{ToneM}_{\texttt{swu} \to \texttt{wu}}^{\texttt{+r}}$. The t-SNE separation of $Z_{\mathrm{IPU}}$, $Z_{\mathrm{WRD}}$, and $Z_{\mathrm{SYL}}$ across both speaking styles suggests that boundary-dependent state resets promote structured learning of level-specific contextual tonal contours, although t-SNE \cite{van2008tsne} alone does not establish improved discriminability.

\section{Conclusion}

We propose ToneMamba, a boundary-aware SSM that models hierarchical linguistic structure through boundary-dependent state resets instead of static embeddings. Combining multi-level resets, shared-weight bidirectional fusion, and rhythm-conditioned gating, ToneMamba learns contextual tonal contours and improves recognition, particularly in conversational speech. The results support state resetting as a flexible approach to modeling hierarchical boundaries. Future work will integrate attention mechanisms and extend ToneMamba to atypical tone recognition for second-language learners.

\section{Acknowledgment}

This work was financially supported by the National Science and Technology Council, Taiwan, under project [115-2221-E-035-070], awarded to the second author. The authors have no relevant financial or nonfinancial interests to disclose. OpenAI Codex was used to assist with English grammar refinement throughout the manuscript and and modifications to the implementation code. The authors reviewed and verified all AI-assisted outputs and take full responsibility for the manuscript and implementation.

% BiBTeX files. The IEEEbib.bst bibliography style file from IEEE produces
% unsorted bibliography lists.
% -------------------------------------------------------------------------
\bibliographystyle{IEEEbib}
\bibliography{strings,refs}

\end{document}